%% file: root.tex
\documentclass[journal,twoside]{IEEEtran}
\IEEEoverridecommandlockouts   

\usepackage{cite}
\usepackage{url}
\usepackage{graphicx}
\usepackage{array,tabularx}
\usepackage{multirow}
\usepackage{bm}
\usepackage{balance}
\usepackage{amsmath,amsfonts}
\usepackage{algorithmic}
\usepackage{array}
\usepackage[caption=false,font=normalsize,labelfont=sf,textfont=sf]{subfig}
\usepackage{textcomp}
\usepackage{stfloats}
\usepackage{url}
\usepackage{verbatim}
\usepackage{makecell}
\usepackage{wasysym}
\usepackage{gensymb}
\usepackage{color}
\newcommand{\charlie}[1]{\textbf{\textcolor[rgb]{0.80,0.00,0.00}{[Charlie::~#1]}}}

\usepackage{color}
\usepackage{ulem}
\newcommand{\rev}[2]{#2}
\newcommand{\revJul}[2]{{#2}}

\begin{document}
\title{Optimizing Out-of-Plane Stiffness for Soft Grippers}
\author{Renbo Su,~\IEEEmembership{Student Member,~IEEE},~Yingjun Tian,~Mingwei Du~and~Charlie C. L. Wang$^{\dagger}$,~\IEEEmembership{Senior Member,~IEEE}
\thanks{
This work was partially supported by the chair professor fund of C.C.L. Wang at the University of Manchester. }
\thanks{All authors are with Department of Mechanical, Aerospace, and Civil Engineering, The University of Manchester, United Kingdom
}%
\thanks{$^{\dagger}$Corresponding Author:~{\tt\footnotesize changling.wang@manchester.ac.uk}}
\thanks{Digital Object Identifier (DOI): see top of this page.}
}

\maketitle

\begin{abstract}
In this paper, we presented a data-driven framework to optimize the out-of-plane stiffness for soft grippers \rev{so that they can have}{to achieve} \rev{the}{}mechanical properties as hard-to-twist and easy-to-bend. The effectiveness of this method is demonstrated \rev{}{i}n \rev{a}{the} design of \rev{}{a} soft pneumatic bending actuator (SPBA). First, a new objective function is defined to quantitatively evaluate the out-of-plane stiffness as well as the bending performance. Then, sensitivity analysis is conducted on the parametric model of a\rev{}{n} SPBA design to determine the optimized design parameters with the help of Finite Element Analysis (FEA). To enable the computation of numerical optimization, a data-driven approach is employed to learn a cost function that directly represents the out-of-plane stiffness as a differentiable function of the design variables. \rev{Gradient-based}{A gradient-based} method is used to maximize the out-of-plane stiffness of \rev{}{the} SPBA while ensuring specific bending performance. \rev{Effectiveness}{The effectiveness} of our method has been demonstrated in physical experiments taken on 3D-printed grippers. 
\end{abstract}

\input{sections/1_Introduction}
\input{sections/2_Related_Work}
\input{sections/3_Problem_Formulation}

\input{sections/4_Data-Driven_Optimization}

\input{sections/5_Experimental_Test_and_Validation}
\input{sections/6_Conclusions_and_Discussion}


\bibliographystyle{IEEEtran}
\bibliography{referencePaper}

\newpage
\section*{Appendix}
\input{sections/APPENDIX}

\end{document}

%% file: sections/1_Introduction.tex
\section{Introduction}
\rev{}{The} Soft gripper\rev{s}{} has caught \rev{a lot of}{much} attention due to \rev{their}{its} good adaptability to the shape of objects to be manipulated. However, the high flexibility of materials (with low Young's modulus) and structures (with low stiffness) used on soft grippers can also lead to undesired out-of-plane deformation, which reduces grasping stability~\cite{scharff2019reducing}. To overcome this deficiency, some researchers added long fiber winding or auxiliary structure\rev{}{s} on soft actuators to enhance the stiffness in different aspects~\cite{galloway2013mechanically,connolly2015mechanical,chen2021soft,scharff2019reducing}. The geometry of a soft gripper design was also optimized to reduce the maximal strain (therefore material failure) in \cite{dammer2018design}. Out-of-plane deformation was not considered in their approach. Moreover, including physical simulation by finite
element analysis (FEA) in the loop of numerical optimization is prone to the problem of robustness. Specifically, \rev{FEA computation conducted on intermediate designs generated during optimization sometime cannot converge caused by }{the optimization process may fail to converge because of} the numerical instability of nonlinear FEA. 
Differently, we present a data-drive\rev{}{n} approach in this paper that shifts the finite element (FE) computation to the step of dataset preparation. This strategy significantly improves the robustness of our method. 

\begin{figure}[t]
\centering
\includegraphics[width=\linewidth]{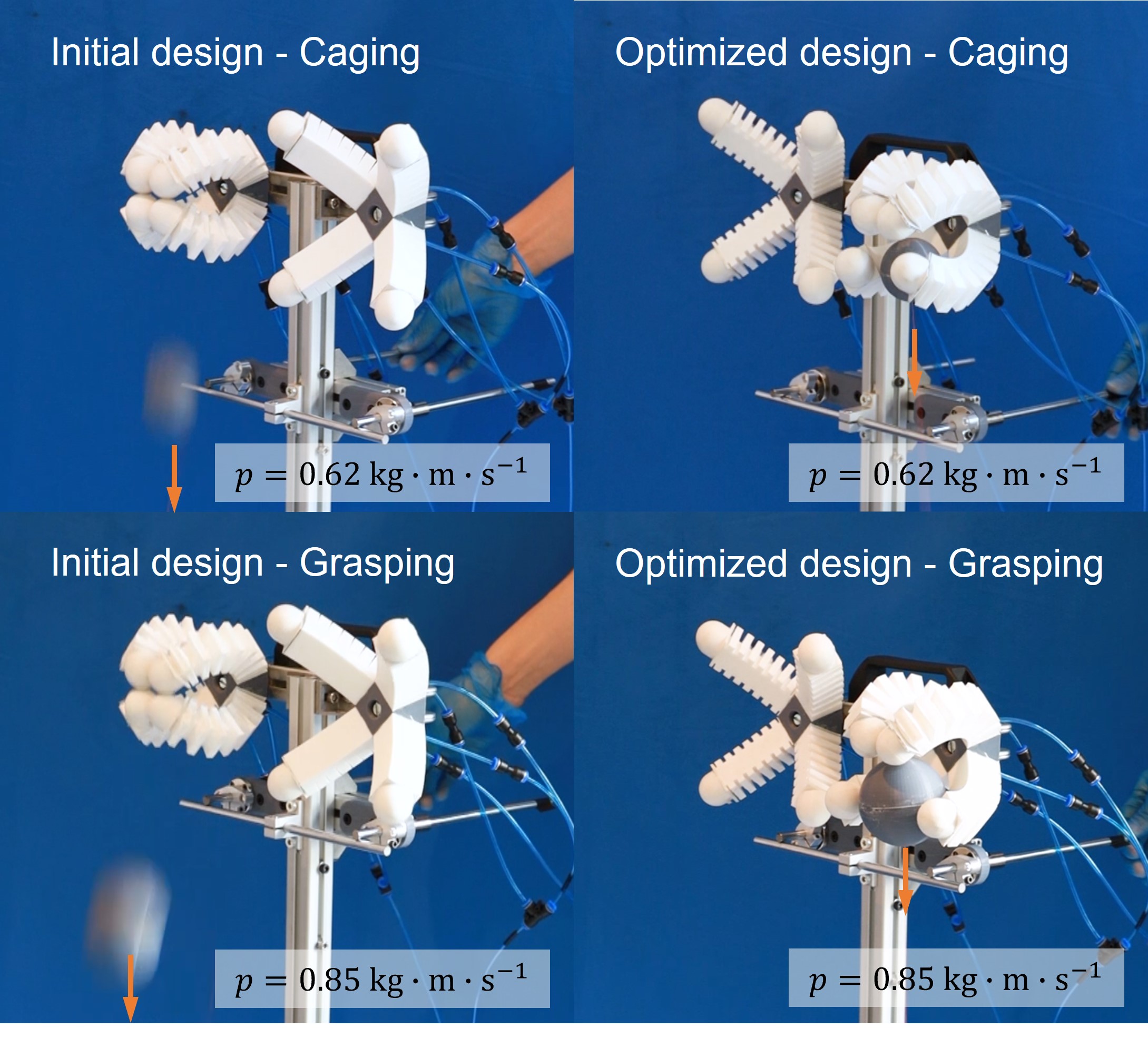}\\
\vspace{-10pt}
\caption{Compared to the initial design of a soft gripper (left), the out-of-plane stiffness has been significantly strengthened on an optimized design (right) \revJul{}{in the dynamic experiment}. When applying the same momentum as side-load (see the arrow as loading direction) to the caged\rev{ / }{/}grasped sphere, the optimized soft gripper can still hold the sphere stably while the sphere escapes away from the initial design.}
\label{fig:cover}
\vspace{-10pt}
\end{figure}

To optimize the out-of-plane stiffness for soft grippers, we define an objective function that can quantitatively capture the requirement to enhance the out-of-plane stiffness via the bending results of a soft actuator when contacting an asymmetric obstacle. With the help of FEA, we conduct sensitivity analysis to effectively reduce the dimension of variables on the parametric model of a \rev{SPBA design}{soft pneumatic bending actuator (SPBA)}. A data-driven optimization framework is proposed in this paper to fit the selected design variables and their corresponding performance defined by the objective function to a differentiable hypersurface in a high-dimensional space. The hypersurface can predict the performance of different geometry in the design space, and therefore can also be employed to find an optimal design by the gradient-based optimization solver. We can effectively maximize the out-of-plane stiffness on \rev{a}{the} SPBA design while keeping a required bending behavior and applying the same air pressure as pneumatic actuation. Physical experiments have been tested to verify the effectiveness of our method under different loads (see Fig.\ref{fig:cover} for an example). 

Our work makes the following technical contributions.
\begin{itemize}
\item An objective function to quantitatively evaluate the out-of-plane stiffness under a fixed bending requirement;

\item A data-driven framework for optimizing the out-of-plane stiffness on parameterized gripper models by learning a differentiable function;

\item A novel design of physical experiment that can apply repeatable loads on grasped\rev{ / }{/}caged objects \rev{for the study of}{to study} grasping stability on soft grippers.
\end{itemize}
\rev{An}{The} SPBA design is employed to demonstrate the functionality of our method. However, the methodology presented here can be generally applied to different designs of soft grippers. 

%% file: sections/2_Related_Work.tex
\section{Related Work}
\subsection{Out-of-plane deformation}
Although grippers using soft-bending actuators have received a lot of attention and been widely studied since the work in \cite{ilievski2011soft}, few people have conducted in-depth research on the failure of soft grippers.
Scharff et al. \cite{scharff2019reducing} were the first to realize that the deformation outward \rev{}{from} the bending plane under asymmetric load is a\rev{n}{} critical issue for unexpected buckling and twisting, which is similar to the two patterns of grasping failure of spherical objects as reported in \cite{brown2021design}. By increasing the torsional constant and adding a stiffening structure, Scharff et al. \cite{scharff2019reducing} reduced the out-of-plane deformation and improved the grip performance. 

Soft-rigid hybrid actuators could be an alternative strategy to solve the problem of out-of-plane deformation. 
In nature, the exoskeletons of crustaceans such as lobster also have similar structures, and hybrid actuators inspired by it have been designed and studied \rev{}{in} \cite{chen2017lobster,chen2017soft,chen2020lobster}, although the authors may not be aware of the effect of this hinge-based structure on the out-of-plane stiffness. In contrast, Lotfiani et al. \cite{lotfiani2020torsional} embedded hinges-based skeleton during \rev{}{the} casting of soft actuators to improve their torsional stiffness. \rev{}{Morrow et al. \cite{morrow2016improving} used 3D printed nails (fingertips) to avoid out-of-plane buckling.}

Torsion coupled with bending is not always detrimental. Some researchers have achieved controllable twist\rev{}{s} through material or structural design\rev{,}{} and obtained winding grippers like vines and tentacles (ref.~\cite{wang2016plant,wang2018programmable}). \rev{}{The joint locking mechanism formed by the conductive PLA bottom layer was added by Al-Rubaiai et al. \cite{al2019soft} to \rev{indirectly}{}limit twisting for soft fingers \rev{}{indirectly}.}
Different from these existing approaches, we propose a method to reduce the out-of-plane deformation on soft grippers by optimizing the parameters of designs.

\subsection{Design\rev{}{ing} the mechanical properties of soft robots}
The mechanical properties of soft robots mainly depend on their geometry and materials, so the desired characteristics are always designed from these two aspects. \revJul{}{A comprehensive review can be found in \cite{chen2020design}.} Here we mainly focus on soft pneumatic robots. 

Morin et al. \cite{morin2014elastomeric} found that the distribution of materials resulted in a high degree of freedom of programmability of the expansion or collapse pattern in their research of elastic cubes made of two soft materials with different stiffness (i.e., Ecoflex and PDMS). Using the same combination of materials, Forte et al. \cite{forte2022inverse} recently proposed a machine learning-based framework for the inverse design of the soft membrane that can achieve the pre-programmed deformation under inflation. Dammer et al. \cite{dammer2018design} demonstrated a 3D-printable linear actuator and optimized the geometry parameters to extend its lifetime by considering the maximal strain. In this paper, we propose to optimize the out-of-plane stiffness by changing \rev{}{the} soft gripper's geometry. 

The bending actuator\rev{}{s} are often designed \rev{by }{based on} the mutual squeeze of two adjacent air chambers after inflation. In the research of Mosadegh et al. \cite{mosadegh2014pneumatic}, the significant differences between the slow pneu-net (sPN) and fast pneu-net (fPN) proposed were  reflected not only in the response speed but also in the actuation pressure. This pioneer\rev{}{ing} work has inspired a lot of new designs for the soft pneumatic bending actuator\rev{s}{~(SPBA)}. For example, Altelbani et al. \cite{altelbani2021design} combined 3D-printable fPN with different bending properties in various ways for powerful structural\rev{-}{} programmability. Based on sPN, Ke et al. \cite{ke2021stiffness} achieved pre-programmed conformal grasping by varying the thickness of the stretchable top layer corresponding to the curvature of different regions of the grasped object. We employ a\rev{}{n} fPN-based design to demonstrate the effectiveness of our approach.

\subsection{Metamodeling for design optimization}
The data-driven approach developed in this paper adopts a strategy similar to the metamodeling research\rev{es}{} in the area of engineering design. As the abstraction of a simulation model, metamodels can effectively reduce computational or experimental cost\rev{}{s} by analytically express\rev{}{ing} the behavior of the tested model \cite{liu2018survey}. In addition to being surrogates for simulation models, differentiable analytical expressions also \rev{make }{enable} metamodel \rev{}{to be} widely used in design optimization. Over the past few decades, \rev{a number of}{several} techniques \cite{cressie1988spatial, rasmussen2003gaussian, dyn1986numerical, clarke2005analysis} have been proposed for metamodeling, which interpolate\rev{}{s} or fit\rev{}{s} a function through some result data and predict\rev{}{s} the new results without additional simulation computing. Theoretically\rev{}{,} a fairly accurate metamodel can be obtained as long as the \rev{dataset of training}{training dataset} is sufficient. A review by Wang and Shan \cite{wang2007review} summarizes five advantages of metamodel-based design optimization (MBDO), including 1) easy to interact with commercial simulators, 2) suitable for parallel computing, 3) better noise resistance, 4) excellent in visualizing the entire design space, 5) robust to error in simulation. Our data-driven framework also share\rev{}{s} all these advantages. 

Park and Dang \cite{park2010structural} proposed a design optimization framework that integrates commercial CAD and CAE package\rev{}{s} to enable a highly automated and fast product development process, which reduced calculation costs. Three cases of design problem\rev{}{s} with the dimensional ranging from low to high demonstrate the advantages of the framework in terms of feasibility, convenience and computational efficiency, etc. A similar approach was used in \rev{the study of}{studying} the collision performance of thin-walled conical tubes by Acar et al. \cite{acar2011multi}; however, the too sparse training points made the metamodel unable to obtain the optimal solution. Using \rev{the }{a} similar strategy, we develop a data-driven framework to optimize the design of soft robots, which minimizes the out-of-plane deformation by finding the `best' geometric parameters of \rev{soft pneumatic bending actuator (SPBA)}{the SPBA}. 

%% file: sections/3_Problem_Formulation.tex
\section{Problem Formulation}
\subsection{Definition of out-of-plane stiffness}
\begin{figure}[t]
\centering
\includegraphics[width=1\linewidth]{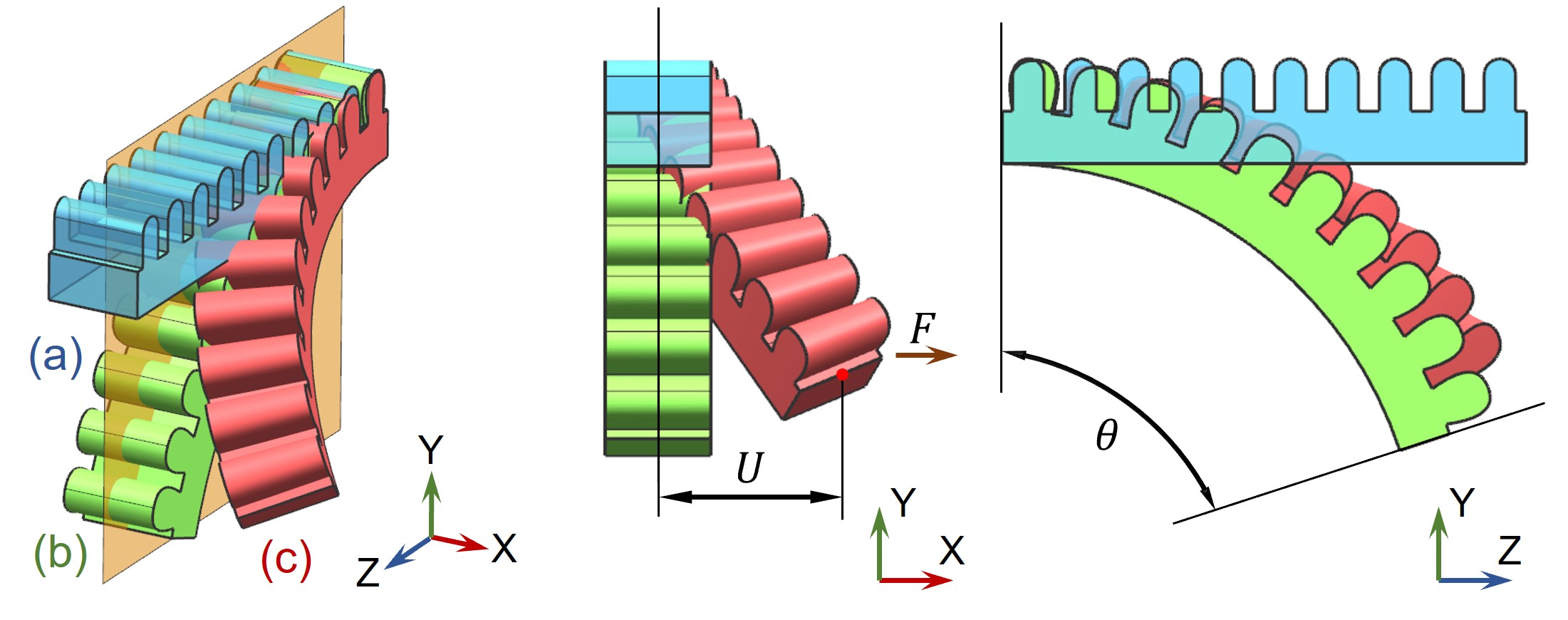}\\
\vspace{-5pt}
\caption{\rev{A}{The} SPBA deforms from \rev{}{(a)} the original state \rev{(a)}{}to \rev{}{(b)} a state with only in-plane bending \rev{(b)}{}under actuation, and exhibits \rev{}{(c)} the out-of-plane deformation \rev{(c)}{}with asymmetric loads when colliding an obstacle with asymmetric shape.}
\label{fig:OoPDefine}
\vspace{-10pt}
\end{figure}

The problem to be solved here is to optimize the out-of-plane stiffness of soft grippers \rev{so that they}{to} become hard to sustain under the out-of-plane deformation while still keep\rev{}{ing} \rev{a}{}relative\rev{}{ly} high flexibility for pure in-plane bending. To quantitatively address the problem of out-of-plane deformation, we define the stiffness for the out-of-plane deformation by presenting an obstacle in \rev{}{a} simple shape that passively generates asymmetric loads to a soft gripper. An actuated SPBA is used as an example to illustrate the definition (see also Fig.\ref{fig:OoPDefine}).

When the SPBA (see Fig.\ref{fig:OoPDefine}(a)) is actuated and starts to collide \rev{}{with} an asymmetric obstacle, a hybrid deformation (see Fig.\ref{fig:OoPDefine}(c)) composed of in-plane and \rev{sideways }{lateral} bending as well as torsion will be generated instead of the pure in-plane deformation (Fig.\ref{fig:OoPDefine}(b)). The \textit{out-of-plane displacement} $U$ is defined as the distance from the center of the SPBA's tip to the pure bending plane. The component of the force \rev{}{perpendicular to the bending plane} generated by the obstacle on the SPBA \rev{that is perpendicular to the bending plane}{} is named \rev{as}{}the \textit{out-of-plane deformation force}\rev{and}{,} denoted by $F$ in the rest of this paper.

The \textit{out-of-plane stiffness} is defined by using $U$ and $F$ as
\begin{equation}
\small\label{eq:OoPDefine}
k_{o}=\frac{F}{U+\varepsilon}
\end{equation}
where $\varepsilon$ is an auxiliary parameter that prevent\rev{}{s} abnormal values caused by \rev{very}{}extremely small denominators. $\varepsilon=1 \mathrm{mm}$ is used in all our tests. Inspired by the previous study~\cite{scharff2019reducing}, a ramp is employed in our work as the asymmetric obstacle. The design of \rev{}{the} ramp should allow \rev{to generate}{generating} a collision force that has the components out of the bending plane (i.e., $F$). 
The \textit{unobstructed bending angle} $\theta$ of \rev{a}{the} SPBA under the same pneumatic actuation is used to represent its bending performance. Note that the exact pneumatic actuation and the position of \rev{}{the} ramp will be adjusted according to the dimensions of \rev{}{the} SPBA and the required unobstructed bending angle $\theta$.

\subsection{Parametric model for optimization}\label{subsecParaModel}
\begin{figure}[t]
\centering
\includegraphics[width=1\linewidth]{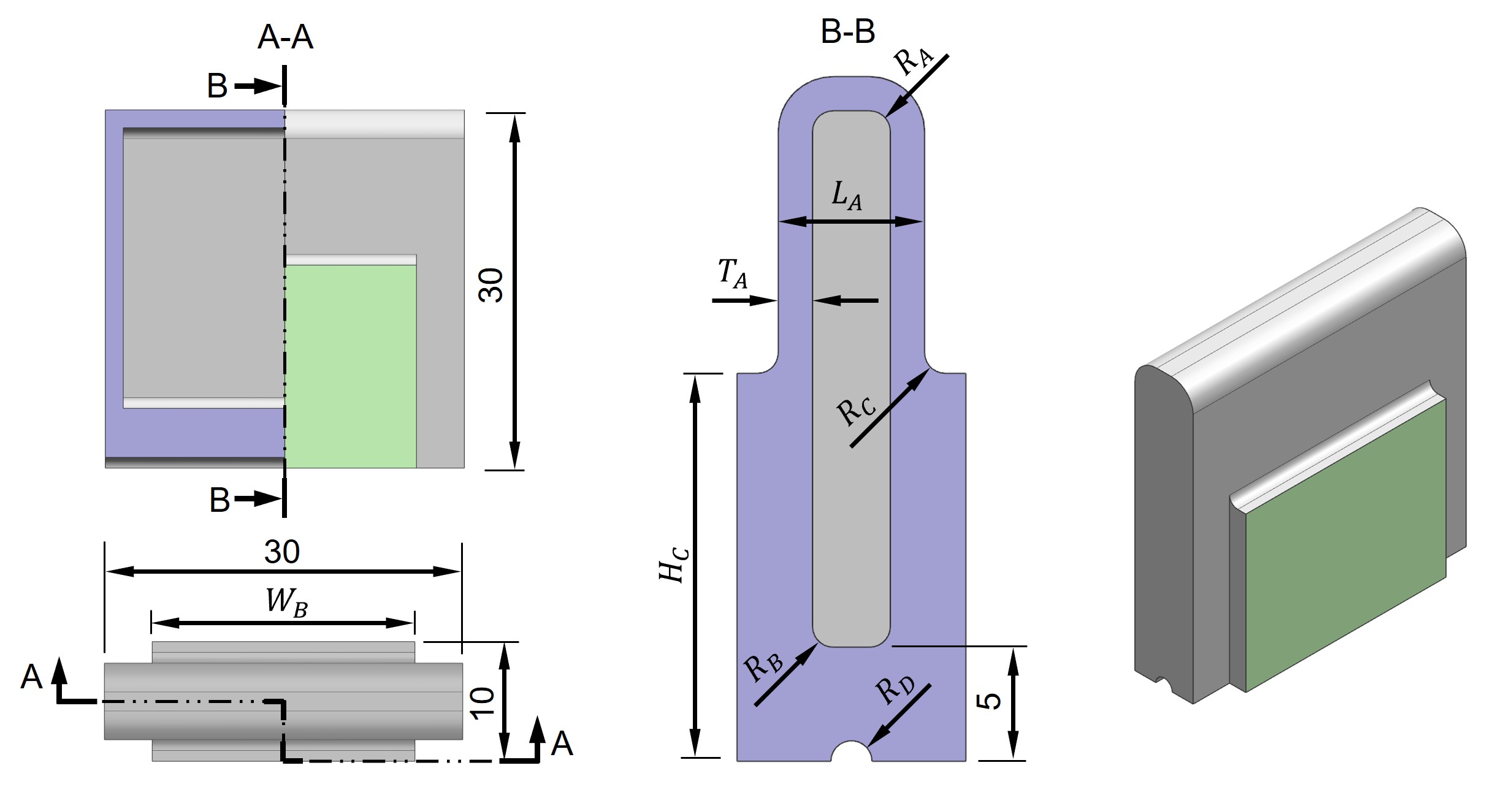}\\
\footnotesize
\vspace{5pt}
\begin{tabular}{l|c|c}
\hline
\textbf{Parameter} &
\textbf{Symbol} & 
\textbf{Range (Unit: mm)}\\
\hline
\hline
Height of the connecting section &
$H_{C}$ & $17.0\pm 8.0$ \\
\hline
Outer length of the air-chamber &
$L_{A}$ & $6.4\pm 3.6$ \\
\hline
Fillet radius of the ceiling &
$R_{A}$ & $0.9 \pm 0.9$ \\
\hline
Fillet radius of the floor &
$R_{B}$ & $0.9\pm 0.9$ \\
\hline
Fillet radius on the connecting section &
$R_{C}$ & $0.9\pm 0.9$ \\
\hline
Radius of the bottom groove &
$R_{D}$ & $0.9\pm 0.9$ \\
\hline
Thickness of the chamber wall &
$T_{A}$ & $1.5\pm 0.5$ \\
\hline
Width of the connecting section &
$W_{B}$ & $22.0\pm 8.0$ \\
\hline
\end{tabular}
\caption{\rev{Parametric}{The parametric} model for the chamber of \rev{a}{the} SPBA, where the range of each parameter is given in the table.}
\label{fig:8ParaMod}
\vspace{-10pt}
\end{figure}

In this paper, an SPBA based on fPN is employed to demonstrate how the optimization is conducted on a parametric design. The SPBA consists of multiple repeated units having the same design, where each unit has its width, height and length as $30 \mathrm{mm} \times 30 \mathrm{mm} \times 10 \mathrm{mm}$. \rev{Different from}{Unlike} the traditional fPN design that is widely used, the width dimension of the connecting block (highlighted in green color in Fig.\ref{fig:8ParaMod}) is allowed to be different from the unit's width. As a result, a gap can be formed between two neighboring units\rev{}{,} and the influence of the gap on stiffness can be controlled by the design parameters $W_{B}$ and $H_{C}$. 
The thickness of the bottom layer is fixed as $5 \mathrm{mm}$ in our design. 
To enlarge the bending flexibility, semicircular grooves with variable radius $R_{D}$ are added below each \rev{air-chamber}{air chamber}. Also, to understand the influence of fillet \rev{radius}{size} on \rev{}{the} SPBA, round corners with variable \rev{radius were}{radii are} added to the inner air chambers and the top of the connecting sections (i.e., $R_{A}$, $R_{B}$\rev{}{,} and $R_{C}$). Eight design parameters contained in the initial design are shown in \rev{the}{}Fig. \ref{fig:8ParaMod}, and the range\rev{}{s} of these parameters are given in the table.

%% file: sections/4_Data-Driven_Optimization.tex
\section{Data-Driven Optimization}
\begin{figure}[t]
\centering
\includegraphics[width=1\linewidth]{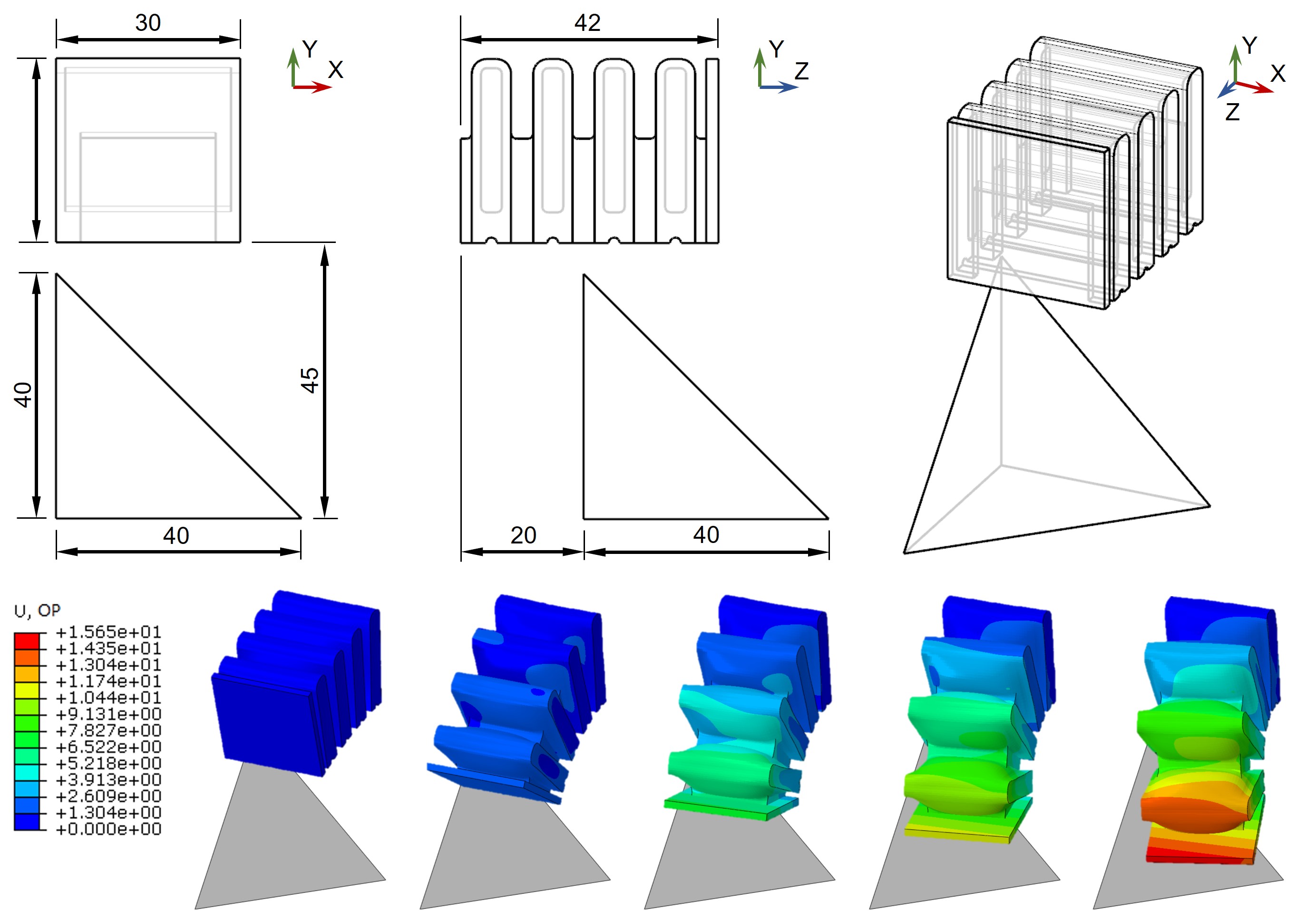}\\
\vspace{-5pt}
\caption{Dimensions and relative positions of SPBA and the \rev{slope}{ramp} as \rev{}{the} obstacle (top). Progressive results of \rev{}{the} simulation are shown \rev{at}{in} the bottom row, where the out-of-plane displacements of points are visualized by colors. \rev{}{The ramp's orientation is selected to be more similar to the grasping scenario.}}
\label{fig:FEAMod}
\end{figure}

%
We proposed a data-driven approach to optimize the out-of-plane stiffness, in which FE simulations of soft grippers under different asymmetric loads are conducted to generate the dataset for learning. To avoid a learning process that is too time-consuming, we need to control the complexity of the model involved in FE simulation. Therefore, an SPBA with four aforementioned units (as a parametric model) \rev{is used to reduce}{reduces} the computational load while ensuring \rev{the}{}sufficient \rev{of its}{}generality of \rev{}{the} SPBA with more units. The dimensions of the SPBA and the \rev{location of the ramp}{ramp's location} are illustrated in Fig.\ref{fig:FEAMod}. Our FE simulations are taken on Abaqus/CAE \rev{by}{}using hexahedral meshes and a frictionless contact model. NinjaFlex is used as the material of \rev{}{the} SPBA in this study and the Yeoh hyperelastic model in \cite{yap2016high} is adopted \rev{}{for simulation}.

\subsection{Sensitivity analysis and model reduction}\label{subsecSensitivity}
Directly meshing the parametric model introduced in Section \ref{subsecParaModel} and sampling the design space of \rev{its }{}all eight parameters is impractical due to \rev{the}{a} large number of elements in the resultant mesh and the large number of samples needed for learning. Sensitivity analysis and model reduction are conducted before generating datasets for learning. 

A problem first noticed was the complex meshes caused by the geometry of \rev{}{the} air channel, which also made it very challenging to generate a full hexahedral mesh for \rev{}{the} SPBA. If the air channel is ignored, the model can be simulated by fewer hexahedral elements\rev{}{,} thus \rev{reduce}{reducing} the computing load of \rev{}{the} simulation. 
Our test results show \rev{that there is}{}no significant difference between the simulation results with vs. without the air channel. In \rev{}{the} free bending simulation, the maximum displacement error is only $2.598\%$ on the SPBA with \rev{4}{four} units. After presenting the obstacle, the error of bending simulation is only $0.996\%$. Analytically, removing the \rev{air-channel}{air channel} from the simulation can slightly increase the bending and torsion resistance of the connecting section -- i.e., the level of deformation is reduced. However, we argue that the observed error is within an acceptable range\rev{}{,} and this reduction can significantly improve the \rev{}{learning} efficiency\rev{of learning}{}. 
\begin{figure}[t]
\centering
\vspace{-10pt}
\includegraphics[width=1\linewidth]{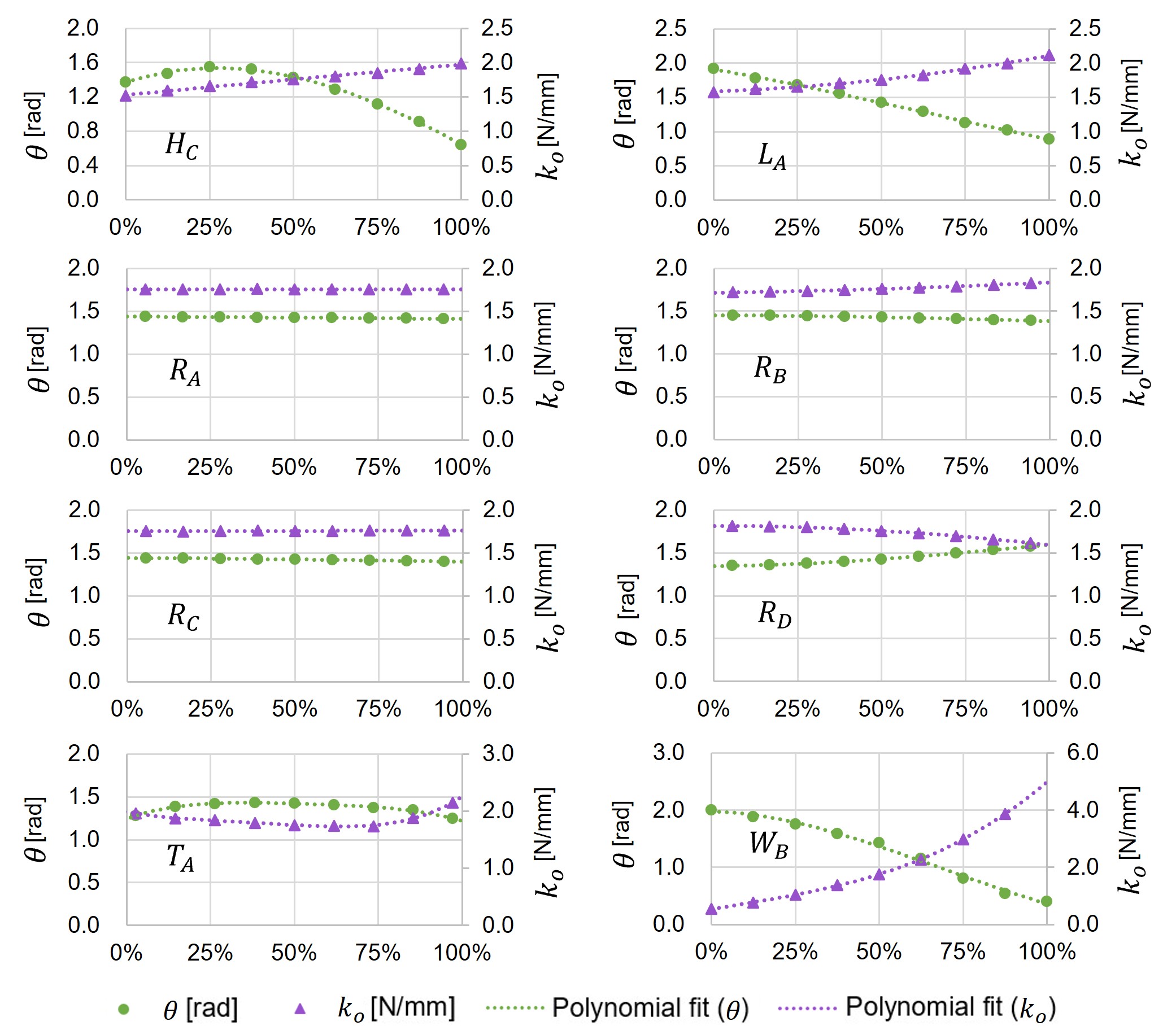}\\
\vspace{-5pt}
\caption{Study the change of of the unobstructed bending angle $\theta$ (i.e., without presenting the obstacle) and the out-of-plane stiffness $k_{o}$ respect to eight parameters of the parametric model presented in Fig.\ref{fig:8ParaMod}. The pneumatic actuation with air pressure at 100kPa is conducted in the simulations on the models generated by different design parameters, where the percentage value is based on the range of each parameter in Fig.\ref{fig:8ParaMod}.}
\label{fig:Sens8Para}
\vspace{-10pt}
\end{figure}

To further improve the efficiency of our data-driven optimization framework, sensitivity analysis is conducted on the eight parameters of the model to reduce the search space to a lower dimension. Only those parameters that can sensitively contribute to the mechanical property of easy-to-bend and hard-to-twist are kept in the learning and optimization routine. Given a fixed air pressure for pneumatic actuation, the easy-to-bend property is studied by the angle of free bending $\theta$ without presenting the obstacle. The \rev{sensitive}{sensitivity} of out-of-plane stiffness $k_o$ (i.e., hard-to-twist) is evaluated\rev{under the same actuation}{} by introducing the asymmetric obstacle \rev{}{under the same actuation}. The whole range of each design parameter is uniformly sampled at 9 points to evaluate how the values of $\theta$ and $k_o$ are influenced by this design parameter. \rev{When changing one parameter, t}{T}he other seven parameters are kept as the median of their corresponding ranges \rev{}{when changing one parameter}. The results are shown in Fig.\ref{fig:Sens8Para}.

\begin{figure}[t]
\centering
\includegraphics[width=1\linewidth]{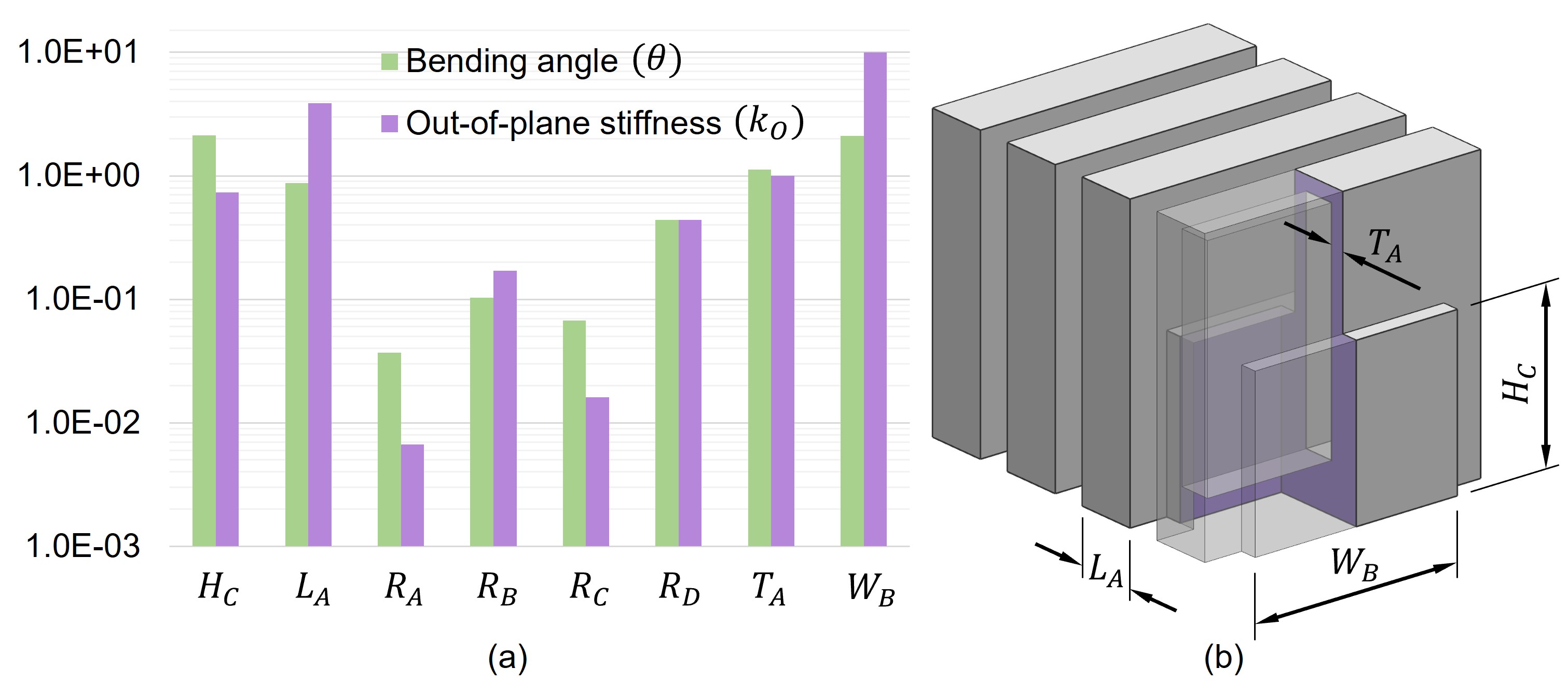}\\
\vspace{-10pt}
\caption{Sensitivity analysis to generate a reduced parametric model: (a) the maximal magnitude of gradients obtained \rev{from}{}on the curves for eight parameters as shown in Fig.\ref{fig:Sens8Para}, and (b) the reduced model with 4 design parameters.}
\label{fig:SensResult}
\vspace{-10pt}
\end{figure}

Polynomial curves are employed to fit the sample points given in Fig.\ref{fig:Sens8Para}, and the maximal magnitude of gradient\rev{s}{} on the curve\rev{s}{} for each design parameter\rev{s}{} \rev{are}{is} considered as the sensitivity of each parameter and plotted as a bar chart in Fig.\ref{fig:SensResult}(a). It can be observed that the fillet parameters $(R_A,~R_B,~R_C)$ \rev{and the size of the bottom groove $(R_D)$}{} have almost no influence on the performance of the designed model. 
\rev{The other }{Although the response of $R_D$ cannot be ignored, we still only retain the} 
four parameters \rev{}{with the highest response} (i.e., $H_C,~L_A,~T_A$ and $W_B$)\rev{that have large value in the sensitivity analysis are retained}{} in the following learning and optimization process. 
\rev{}{This is because the amount of training samples required to learn a 5-dimension design space is much larger than 4-dimensional.}
After this simplification, the parametric model with only four parameters is \rev{as}{}shown in Fig.\ref{fig:SensResult}(b). After removing some parameter\rev{}{s}, the range\rev{}{s} of retained parameters \rev{are}{were} slightly changed into the ranges \rev{as }{}below.
\begin{align}\label{eq:4ParaConstraint}
 8.0 \leq H_{C} \leq 30.0, &   &   2T_{A} < L_{A} < 10.0, \\ 
 1.0 \leq T_{A} \leq 2.0, &  &    14.0 \leq W_{B} \leq 30.0. \nonumber
\end{align}
Note that the values of $T_A$ and $L_A$ \rev{are}{were} constrained according to the model's feasibility.

\subsection{Data-driven modeling}
Gradient-based optimization is employed in our work to determine a design with the `best' parameters that can provide strong out-of-plane stiffness while keeping good flexibility in bending. Directly computing the derivative of $k_o(\bm{x})$ with $\bm{x}=(H_C, L_A, T_A, W_B)$ by numerical difference is prone to the problem of truncation error and the case of not \rev{converge }{converging} in nonlinear FE computing when hyperelastic materials are employed. Differently, we conduct a data-driven approach to learn a set of proxy functions $\Tilde{U}(\bm{x})$, $\Tilde{F}(\bm{x})$ and $\Tilde{\theta}(\bm{x})$ for the out-of-plane displacement $U$, the out-of-plane deformation force $F$ and the unobstructed bending angle $\theta$ respectively. All are represented as continuous functions w.r.t. $\bm{x}$. 

Relatively uniform samples need to be generated in the 4D space of design parameters. The Sobol sequence has become a \rev{favourable }{favorable} tool for quasi-random sampling because of its high dispersion under a small number of samples and good uniformity under high-dimensional problems \cite{sobol2011construction}. We first use the Sobol sequence to generate random points in $\Re^4$ inside a hyper-cube with unit length. Then, each random point is converted to a sample $\bm{x}_i$ ($i=1,\ldots,n$) in the design space of $H_C, L_A, T_A$ and $W_B$ according to their feasible ranges (i.e., Eq.(\ref{eq:4ParaConstraint})). The points that do not satisfy the constraint $2 T_A < L_A$  will be excluded. For each sample $\bm{x}_i$ in the feasible region, two FE simulations with and without presenting the obstacles are conducted with a fixed air pressure to obtain its corresponding values of $U_i$, $F_i$ and $\theta_i$. When the nonlinear FE simulation at a sample point does not converge, we randomly generate another sample around this point and conduct the FE simulation again until it converges. 

After collecting enough \rev{number of sample}{} points $\{\bm{x}\}$ and their corresponding values of $(U_i, F_i, \theta_i)$, a regression model based on the neural network of \textit{radial basis functions} (RBF)~\cite{dyn1986numerical,dinh2002reconstructing,turk2005shape} is computed as
\begin{equation}\label{eq:FullFunc}
\centering
[U_i, F_i, \theta_i]=f(\bm{x}_i) \quad (\forall i=1, \ldots, n).
\end{equation}
Three different functions are learned for $\Tilde{U}(\bm{x})$, $\Tilde{F}(\bm{x})$ and $\Tilde{\theta}(\bm{x})$. For ease of explanation, we take $\Tilde{U}(\bm{x})$ as an example to introduce the fitting method of the RBF network.
The RBF for $U$ can be expressed as a weighted summation equation as
\begin{equation}\label{eq:RBF1}
\centering\small
\Tilde{U}(\bm{x})=P(\bm{x})+
\sum_{j=1}^{n}w_j\phi(\left\|{\bm{x}-\bm{c}_j}\right \|).
\end{equation}
$\phi$ is the basis function, where the 2D thin-plate spline function form \cite{turk2005shape} \rev{was}{is} used in our implementation. That is 
\begin{equation}\label{eq:2dThinPlate}
\centering\small
\phi(r)=r^2\ln r
\end{equation}
$w_j$ is the weight for each basis function. In this formulation, $\bm{c}_j$ means the centre of each basis function, $\left\|{\bm{x}-\bm{c}_j}\right \|$ means the distance between the input argument and each centre, which can also be denoted as $r_j$. Collocation strategy is employed here -- i.e., each sample is employed for the center of a basis function as $\bm{c}_j=\bm{x}_j$. $P(\bm{x})$ is a polynomial term related to the basis function used for affine transformation, which is
\begin{equation}
\label{eq:4dPoly}
\centering
P(\bm{x})=p_0+ (p_1, p_2, p_3, p_4) \cdot \bm{x}.
\end{equation}
The function $f(\bm{x})$ can be determined by imposing the interpolation constraints in Eq.(\ref{eq:FullFunc}) and the compatibility conditions to compute the values of $p_k$s and $w_j$s as $n+5$ unknown variables. $\Tilde{F}(\bm{x})$ and $\Tilde{\theta}(\bm{x})$ can be obtained in the same way.

\begin{figure}[t]
\centering
\includegraphics[width=1\linewidth]{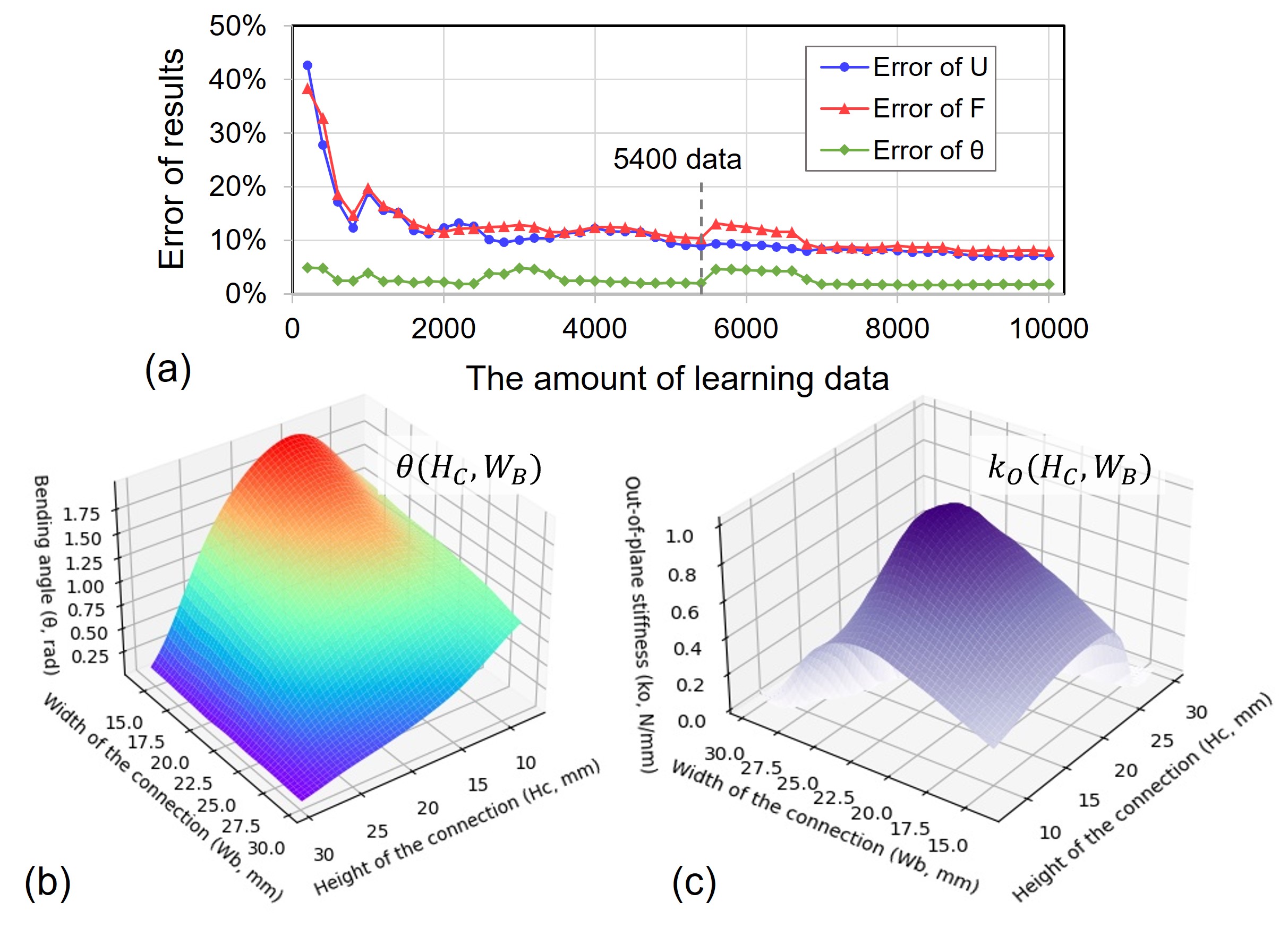}\\
\vspace{-5pt}
\caption{Study of learning error w.r.t. the number of samples in the dataset -- 80\% are used for training and 20\% are used for testing. (a) Learning errors of $U(\bm{x})$, $F(\bm{x})$ and $\theta(\bm{x})$. (b,c) Data-driven model of the bending function $\theta$ and the out-of-plane stiffness function $k_o$ w.r.t. the height $H_C$ and the width $W_B$ of the connecting block  while setting other design parameters as $T_A=1.5$ and $L_A=6.0$.}
\label{fig:Error_Metamod}
\vspace{-10pt}
\end{figure}

To evaluate the accuracy of fitting results, we define the fitting error as $\sigma / \mu$ (named as coefficient of variation). $\sigma$ is the standard deviation of the differences between the estimated values and the real values, and $\mu$ is the mean of real values. 80\% of the dataset is used for training\rev{,}{} and the remaining 20\% is employed for testing. We conducted a study by using different number\rev{}{s} of samples in the training. The results are shown in Fig.\ref{fig:Error_Metamod}(a). It \rev{is}{was} found that \rev{}{the set of} 5400 \rev{samples for learning }{learning samples} (i.e., $5400 \times 80\% = 4480$ samples for fitting) \rev{has}{achieves} a good balance \rev{of}{between} accuracy and efficiency. 
We use the \rev{this}{} model \rev{learned }{trained} by $4480$ samples in the rest of this paper. The unobstructed bending function $\Tilde{\theta}(\bm{x})$ and the out-of-plane stiffness function $\Tilde{k}_0(\bm{x})=\Tilde{F}(\bm{x}) / (\Tilde{U}(\bm{x})+ \epsilon)$ obtained from learning are visualized in Fig.\ref{fig:Error_Metamod}(b) and (c). 

\subsection{Numerical optimization}
After obtaining the differentiable proxy functions $\Tilde{U}(\bm{x})$, $\Tilde{F}(\bm{x})$ and $\Tilde{\theta}(\bm{x})$, we are able to solve the problem as follows to optimize the out-of-plane stiffness of \rev{a}{the} SPBA while keeping its flexibility in unobstructed bending.
\begin{equation}\label{eq:FinalOptm}
\max_{\bm{x}\in\mathcal{F}} \frac{\Tilde{F}(\bm{x})}{\Tilde{U}(\bm{x})+\varepsilon} \qquad s.t., \Tilde{\theta}(\bm{x})=\vartheta
\end{equation}
where $\mathcal{F}$ is the feasible region of \rev{a}{the} SPBA design as constraints defined in Eq.(\ref{eq:4ParaConstraint}). The \rev{other }{}constraint gives the required \rev{bending }{}angle in \rev{radian}{unobstructed bending} as $\vartheta$, which is a value proportional to the number of\rev{chamber}{} units used in \rev{}{the} SPBA. For example, to achieve a bending angle of about $150 ^{\circ}$ with \rev{a bending with}{} 7 units under the pressure of 100kPa, the constraint here should be $\vartheta=1.5 \mathrm{rad} \approx 150^{\circ} \times \frac{4}{7}$.

%% file: sections/5_Experimental_Test_and_Validation.tex
\section{Results and Experimental Verification}
\rev{Based on the optimized design of SPBA unit bellow obtained above, we design a gripper to test and verify the effectiveness of our optimization. The gripper consists of 4 fingers, where each finger is formed by 7 repeated SPBA units together with a hemispherical fingertip. As shown in Fig.\ref{fig:GrpDes_SimRst}(c), the assembly direction of the fingers is 60 degrees from the central axis. The initial design and the optimized design are employed on the same gripper design to conduct comparison by both the numerical simulation and the physical experiment.}{
We first tested the \rev{results of optimization}{optimization results} on a finger with 4 repeated SPBA units, which is consistent with the optimization setup. After that, the experimental tests are conducted on a gripper with four fingers to verify the performance of design optimized by our method.
}

\subsection{Results on 4-unit SPBA}
The sequential least-squares programming is used in our work to solve the optimization problem defined in Eq.(\ref{eq:FinalOptm}). The parameters of the initial guess \rev{}{(by using the median of the design parameter's range)} and the optimized design are listed and compared in Table \ref{tab:4ParaVal}\rev{, where the results in $\theta$ and $k_o$ are generated by running FE simulations again on the optimized design. It can be found that the out-of-plane stiffness has been improved by 88\% while keeping nearly the same flexibility in unobstructed bending.}{, and the performances of the two designs are compared in both FE simulation and physical tests.}

\begin{table}[t]
\caption{The design parameters and results of \rev{a}{the} SPBA model before and after optimization}
\label{tab:4ParaVal}
\vspace{-8pt}
\begin{center}
\begin{tabular}{c|c|c|c}
\hline
~& \textbf{Symbol} & \textbf{Initial design} & \textbf{Optimized design}\\
\hline
\hline
\multirow{4}*{\textbf{Parameter}} &
$H_{C}$ (mm) & 19.0 & 8.814 \\
\cline{2-4}
~& $L_{A}$ (mm) & 6.0 & 9.999 \\
\cline{2-4}
~& $T_{A}$ (mm) & 1.5 & 1.0 \\
\cline{2-4}
~& $W_{B}$ (mm) & 22.0 & 30.0 \\
\hline
\hline
\multirow{2}*{\textbf{\rev{}{FEA} Result}} &
$k_{o}~$(N / mm) & 0.807 & 1.518 \\
\cline{2-4}
~& $\theta~$(rad) & 1.514 & 1.500 \\
\hline
\multirow{2}*{\rev{}{\textbf{Physical Result}}} &
\rev{}{$k_{o}~$(N / mm)} & \rev{}{0.965} & \rev{}{1.689} \\
\cline{2-4}
~& \rev{}{$\theta~$(rad)} & \rev{}{1.492} & \rev{}{1.477} \\
\hline

\end{tabular}
\end{center}
\end{table}

\begin{figure}[t]
\centering
\vspace{-8pt}
\includegraphics[width=1\linewidth]{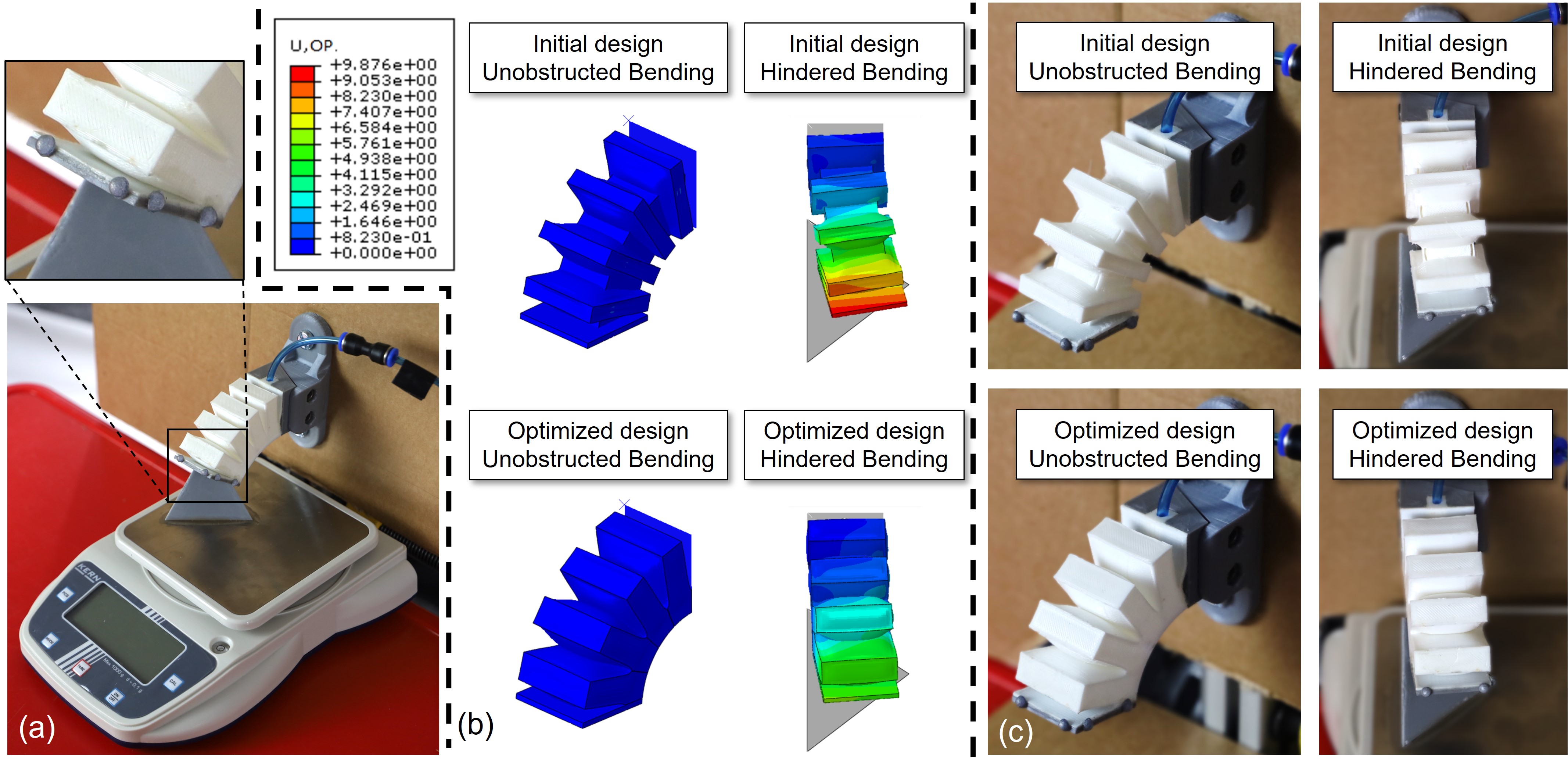}\\
\vspace{-5pt}
\caption{
\rev{}{(a) The hardware setup for verification taken on 4-unit SPBA. 
The results of (b) FE simulation (with out-of-plane displacements being visualized by colors) and (c) physical tests taken on the initial design and the optimized design, where the \revJul{top row gives}{left columns give} the unobstructed bending and the \revJul{bottom row shows}{right columns show} the hindered bending.}
}\label{fig:4Unit}
\vspace{-10pt}
\end{figure}

\rev{}{The material properties and simulation models employed in FEA-based verification are the same as those used in learning and optimization.  
It can be observed from Fig.\ref{fig:4Unit}(b) and Table \ref{tab:4ParaVal} that the out-of-plane stiffness increases by about $88.1\%$ when using the same actuation pressure as $100\mathrm{kPa}$ and precisely keeping the unobstructed bending angle $\vartheta$ as 1.5rad.} 

\rev{}{The physical tests are conducted on 4-unit SPBAs 3D printed by filament deposition using NinjaFlex 85A. The experiment setup is as given in Fig.\ref{fig:4Unit}(a, c), where a ramp is fixed on an electronic scale (KERN EHA1000-1) to measure the out-of-plane deformation force. The out-of-plane displacement and the bending angle are obtained by using a vision system. As a result, about a $75.0\%$ increase can be observed on the out-of-plane stiffness $k_{o}$ (see Table \ref{tab:4ParaVal}) while keeping nearly the same unobstructed bending angle $\theta$. This result is consistent with the FEA-based verification.}

\begin{figure}
\centering
\includegraphics[width=1\linewidth]{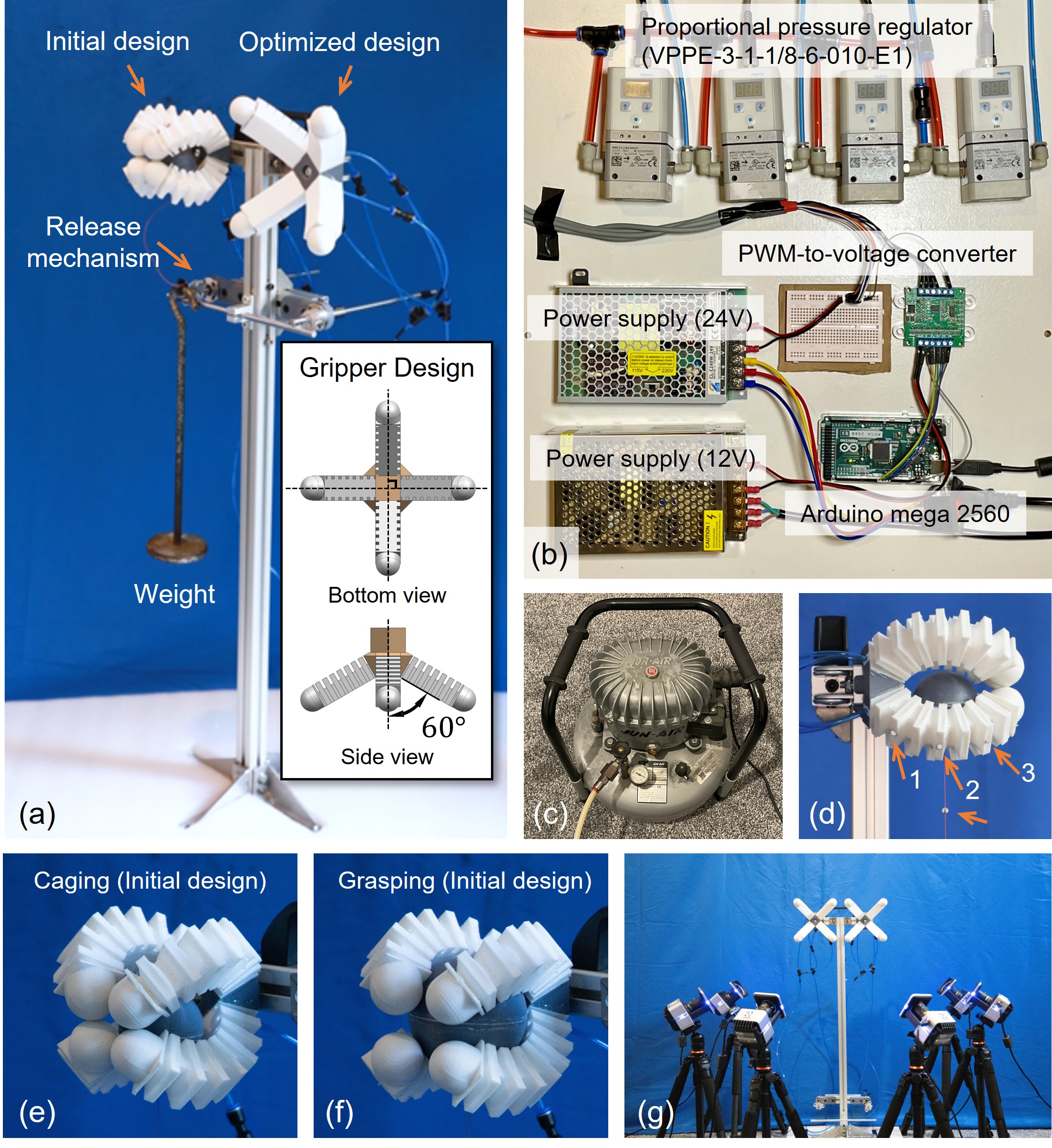}\\
\vspace{-5pt}
\caption{Our hardware setup for physical \rev{}{caging/grasping} experiments: (a) the frame for soft grippers with the release mechanism\rev{}{~and the gripper design}, (b) the control module for air pressure, (c) the Jun-Air Model 6-15 Litre Air compressor, (d) distributed markers on the soft grippers and the pulling string, (e) caging a $\diameter 50 \mathrm{mm}$ sphere, (f) grasping a $\diameter 70 \mathrm{mm}$ sphere, and (g) the motion capture system.}
\label{fig:ExpSet}
\vspace{-10pt}
\end{figure} 

\subsection{Hardware for grasping experiment}
Based on the optimized design of the SPBA unit obtained above, we design\rev{}{ed} a gripper to test and verify the \rev{effectiveness}{generalized performance} of our optimization. The gripper consists of 4 fingers, where each finger is formed by 7 repeated SPBA units \rev{together with}{and} a hemispherical fingertip. As shown in Fig.\ref{fig:ExpSet}(a), the assembly direction of the fingers is 60\degree~from the central axis.\rev{The initial design and the optimized design are employed on the same gripper design to conduct comparison by both the numerical simulation and the physical experiment.}{}

Physical experiments are taken on two grippers (initial design vs. optimized design) \rev{fabricated by filament deposition based 3D printing using}{with 7-unit SPBAs that are 3D printed by} NinjaFlex, where the hemispheres at the tips of 4 fingers are made by PLA. As shown in Fig.\ref{fig:ExpSet}(a), both grippers are mounted on a frame that is able to apply quasi-static loads and dynamic loads (by the designed release mechanism) onto the caged\rev{ / }{/}grasped objects. The air compressor and the pressure control module employed in our hardware setup are shown in Fig.\ref{fig:ExpSet}(b, c).
To measure the out-of-plane displacements of the fingers, a motion capture system with six Vicon Vero v1.3 cameras (Res.: $1280 \times 1024$ pixels) is used to track the 3D positions of markers. The system can provide real-time data streaming up to 250FPS. For each finger, three markers are placed on the top of the 2nd, \rev{the}{}4th, and the 6th bellows (see Fig.\ref{fig:ExpSet}(d)). The bending plane can be determined by the position of these markers under a collision-free bending. The performance of the optimized soft gripper in both caging (Fig.\ref{fig:ExpSet}(e)) and grasping (Fig.\ref{fig:ExpSet}(f)) are tested in our experiments. Note that air pressure is fixed as 100kPa for soft grippers in all experiments reported in this paper.

\begin{figure}[t]
\centering
\includegraphics[width=1\linewidth]{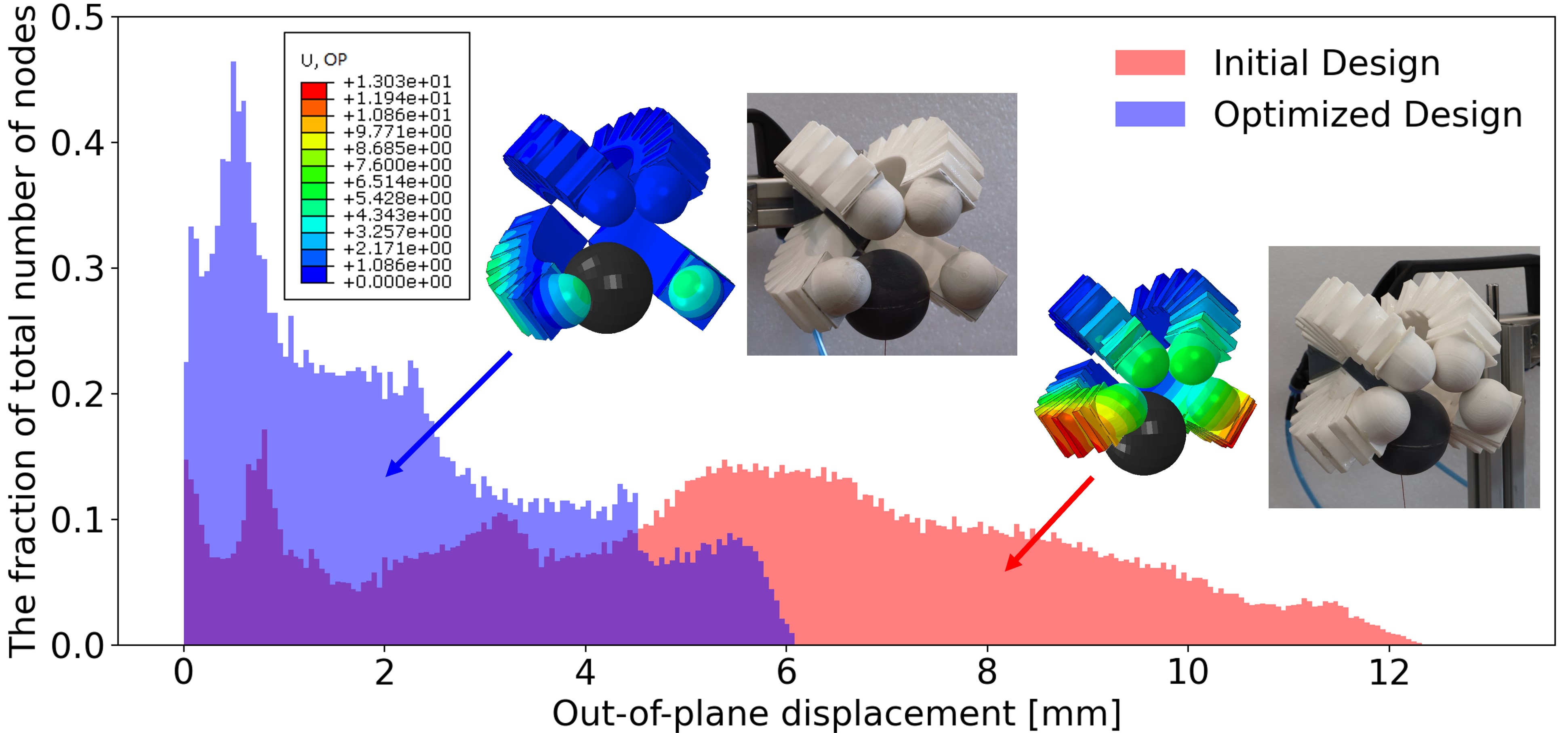}\\
\vspace{-5pt}
\caption{
\rev{A soft gripper with 4 fingers are used to test the performance of our optimized design, where (a) each finger consists of 7 SPBA units as discussed above. (b) A static load is applied to compare the out-of-plane displacements on the initial design vs. the optimized design, which are visualized by the histogram and the colormap in (b).}
{The statistical histogram of out-of-plane displacements (also visualized as contour plots) on all mesh nodes in FE simulations, where the caged spheres are pulled out slowly for 40mm. Photos are the results of physical tests under the same conditions.}
}\label{fig:Quasi_Rst}
\vspace{-3pt}
\end{figure}

\begin{figure}[t]
\centering
\includegraphics[width=1\linewidth]{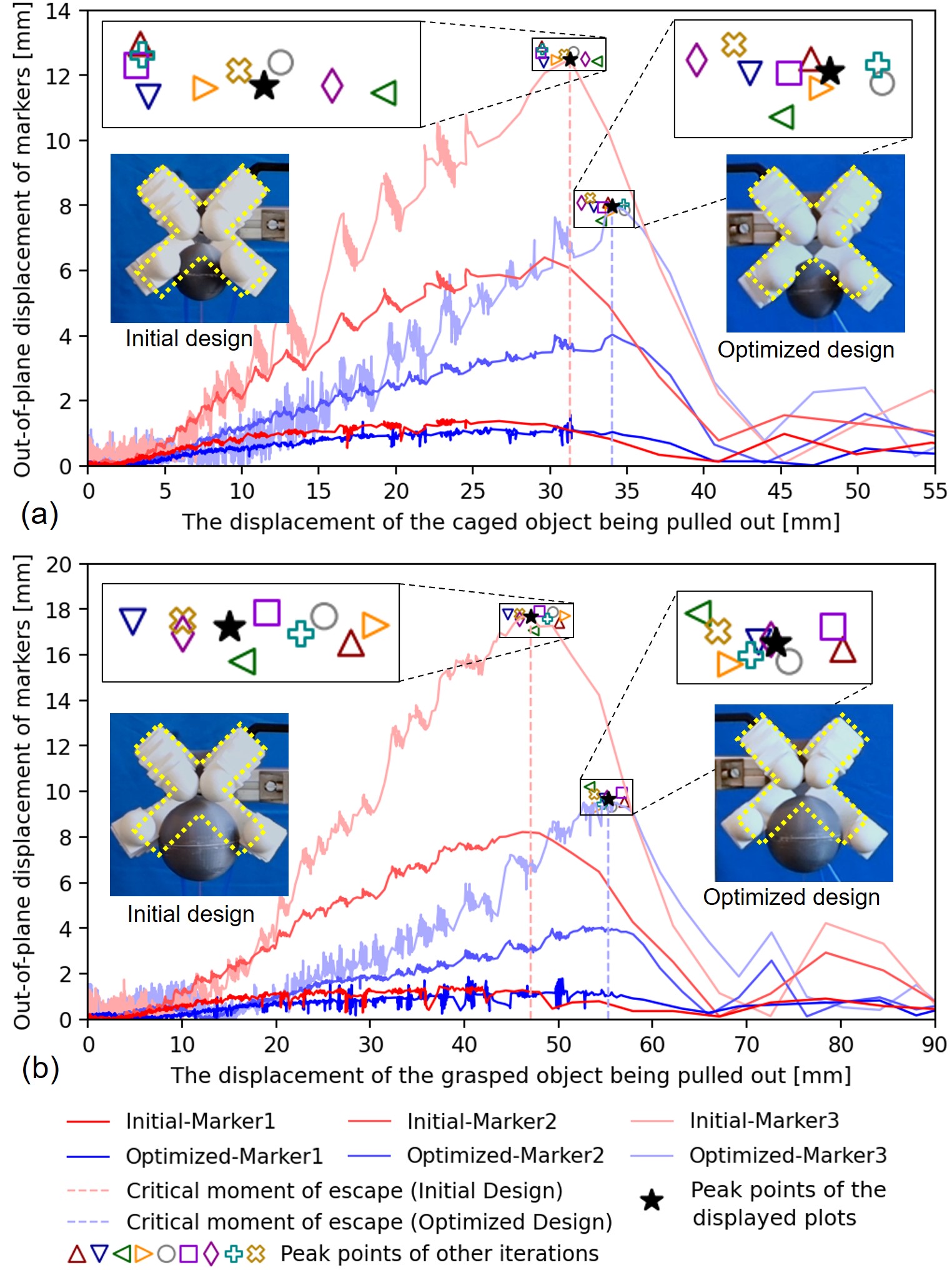}\\
\vspace{-5pt}
\caption{The out-of-plane displacement of markers on both the initial and the optimized designs with respect to the escape displacement of (a) \rev{a}{the caged} sphere and (b) \rev{a}{the} grasped sphere. \rev{}{The shape of a gripper before deformation is given by yellow dash lines.}}
\label{fig:ExpRst1}
\vspace{-10pt}
\end{figure}

\subsection{Quasi-static experiment}
\rev{We first test the soft grippers by applying quasi-static loads.
Specifically, a }{
The experiment aims to compare the out-of-plane stiffness of the gripper fingers on the designs before and after optimization. In order to facilitate the measurement and avoid the influence of inertia on the results, a quasi-static load is applied on a} caged sphere (Diameter: 50mm) \rev{is pulled}{to pull it} out of the gripper slowly through \rev{a }{an inextensible} string.

\rev{}{
In the FE simulation, a displacement of 40mm is applied to the sphere before it escaped from the cage. The results can be found in Fig.\ref{fig:Quasi_Rst}. 
The out-of-plane displacements on all nodes of FEA results are visualized as the colormap and the statistical histograms. It can be observed that the optimized gripper has smaller out-of-plane displacements. This result also demonstrates that the optimization taken on a simplified model with 4 SPBA units can still show its superiority of the out-of-plane stiffness on the gripper with 7 repeated units.}

\rev{The }{In physical tests, the} displacement applied to the caged object is traced by markers attached to the \rev{inextensible}{}string for pulling the object. 
The out-of-plane displacements of the 3 markers on the fingers are \rev{}{also} tracked by the motion capture system. 
\rev{The }{When the ball is pulled out, the} relationship \rev{}{curves} between the caged object's displacement and the out-of-plane displacement \rev{is }{can be} plotted \rev{}{as} in Fig.\ref{fig:ExpRst1}(a). 
\rev{}{The same test was taken 10 times and the peak points from all tests are marked on the graph with different symbols.}
It can be observed that the \rev{}{optimized} gripper \rev{with optimized design}{}will generate less out-of-plane deformation\rev{s}{}, which is especially noticeable on the marker near the \rev{finger tip}{fingertip} (Marker 3) -- i.e., the maximum out-of-plane displacement of the unoptimized gripper \rev{}{, on average,} is about \rev{$1.57\times$ }{$1.59\times$} of the optimized one. If the stage before the Marker 3 reaches the \rev{maximum out-of-plane displacement}{peak} is defined as `stable' caging, we can find that the optimized soft gripper allows a much larger value\rev{s}{} on the object's escape displacement under stable caging. 
\rev{Similar}{A similar} result can be obtained on the test with a \rev{}{larger} sphere (Diameter: 70mm) under grasping -- see Fig.\ref{fig:ExpRst1}(b). For \rev{the markers}{Marker 3, which is} closer to the fingertip\rev{(Marker 3)}{}, the mean of maximum out-of-plane displacement of the unoptimized gripper is about \rev{$1.83\times$}{$1.82\times$} of the optimized one. 
In short, both the caging and the grasping become more \rev{robust}{stable} on \rev{an}{the} optimized gripper. 

\subsection{Dynamic experiment}
\rev{The other}{Another} interesting study \rev{is to investigate}{investigates} the stability of grasping\rev{ / }{/}caging under dynamic loads. Load in the form of instantaneous impact is an important reason for grasp\rev{ing}{} failure. In this experiment, the impact resistance is compared \rev{on}{between} two designs. The grippers are \rev{firstly}{first} actuated to grasp\rev{ / }{/}cage a sphere that is initially connected to a 0.5kg weight by an inextensible string. The weight does not apply any load to the sphere at the beginning of the experiment until we pull the trigger of the release mechanism (see the illustration in Fig.\ref{fig:ExpSet}(a)). When the weight fall\rev{}{s}\rev{down}{} from \rev{different}{various} height\rev{}{s} $h$, different momentum as $p=m\sqrt{2gh}$ will be applied to the grasped\rev{ / }{/}caged sphere with $m=0.5\mathrm{kg}$ being the mass of the weight. We compare the stability of a gripper by checking if the grasped\rev{ / }{/}caged sphere will escape by applying different momentum as impact. Spheres with diameters \rev{}{of} 50mm and 70mm are used for caging and grasping respectively. 

As already shown in Fig.\ref{fig:cover}, we use\rev{}{d} 78mm and 146mm as height $h$ for the caged sphere and the grasped sphere\rev{}{,} respectively\rev{. This}{, which} in fact applie\rev{s}{d} 0.62 and 0.85 (Unit: $\mathrm{kg} \cdot \mathrm{m} / \mathrm{s}$) as impact momentum $p$ to the spheres. With a fixed actuation pressure \rev{as}{of} 100kPa, \rev{}{ten repeated tests showed the same conclusion --} the initial design \rev{was}{is} unable to hold the spheres while the optimized design \rev{could}{can} withstand the impact. The results of \rev{}{the} comparison can also be found in the supplementary video. 

%% file: sections/6_Conclusions_and_Discussion.tex
\section{Conclusions and Discussion}
A data-driven framework is proposed in this paper to optimize the out-of-plane stiffness for soft grippers so that more stable grasping\rev{ / }{/}caging can be realized while keeping their flexibility in bending. In this work, we first developed a new objective function quantitatively evaluate the out-of-plane stiffness under a fixed bending requirement. After that, we introduced a data-driven framework for optimizing the out-of-plane stiffness on parameterized gripper models by learning a differentiable function. Sensitivity analysis has been \rev{conduced}{conducted} on the parametric model of \rev{a}{the} SPBA design to determine the variables to be optimized with the help of Finite Element Analysis (FEA). The optimized design is finally computed by a gradient-based optimizer. The effectiveness of this method is demonstrated \rev{on }{in} \rev{a}{the} design of \rev{}{the} soft pneumatic bending actuator (SPBA). Both FE simulations and physical experiments are conducted on a gripper with 4 fingers \rev{each of which consists of 7 SPBA units}{(each finger consists of 7 SPBA units)} \rev{}{to} verify the performance of our approach. The results are very encouraging as the optimized design outperform\rev{}{s} the initial \rev{design}{one} under both \rev{}{quasi-}static and dynamic loads.


\rev{The results of optimized SPBA show that the gap between two adjacent bellows should be as narrow as possible to maximize the out-of-plane stiffness. However, such a design may lead to a lot of challenges in fabrication and also result in stress concentration. A possible future research is to use multiple materials in the design optimization. }
{The optimization result shows a design with 1) minimized gap -- i.e., an increased value of $L_A$ when keeping the length of each unit fixed and 2) a connection part with wider and shorter cross-section -- i.e., increased $W_B$ and reduced $H_C$. Considering the connection part as a `beam', the principle behind this optimization is that the evolution intends to generate a short and wider `beam' with less height to resist lateral bending and twisting. 
%
\revJul{}{The sensitivity analysis is conducted at the median of parameter's range, which may not cover all possible variations although we did not find any issue by this simplification in our experiment.}
The focus of our work is based on the optimization of structure design when the materials are not allowed to change. However, for the design of actuators without large shape/geometry variation\rev{}{,} such as Galloway design \cite{galloway2013mechanically}, changing the material stiffness is important to enhance the out-of-plane stiffness.
%
Lastly, when a smaller gap (i.e., a larger $L_A$) and a thinner shell (i.e., a smaller $T_A$) are obtained on our optimized design, the bellows \revJul{can be inflated}{inflate} and collide faster. The collision between inflated bellows leads to \revJul{an easy-to-bend result, which means a large}{a larger} bending angle under a given pressure.
}

%% file: sections/APPENDIX.tex

\begin{figure}[h]
\centering
\includegraphics[width=\linewidth]{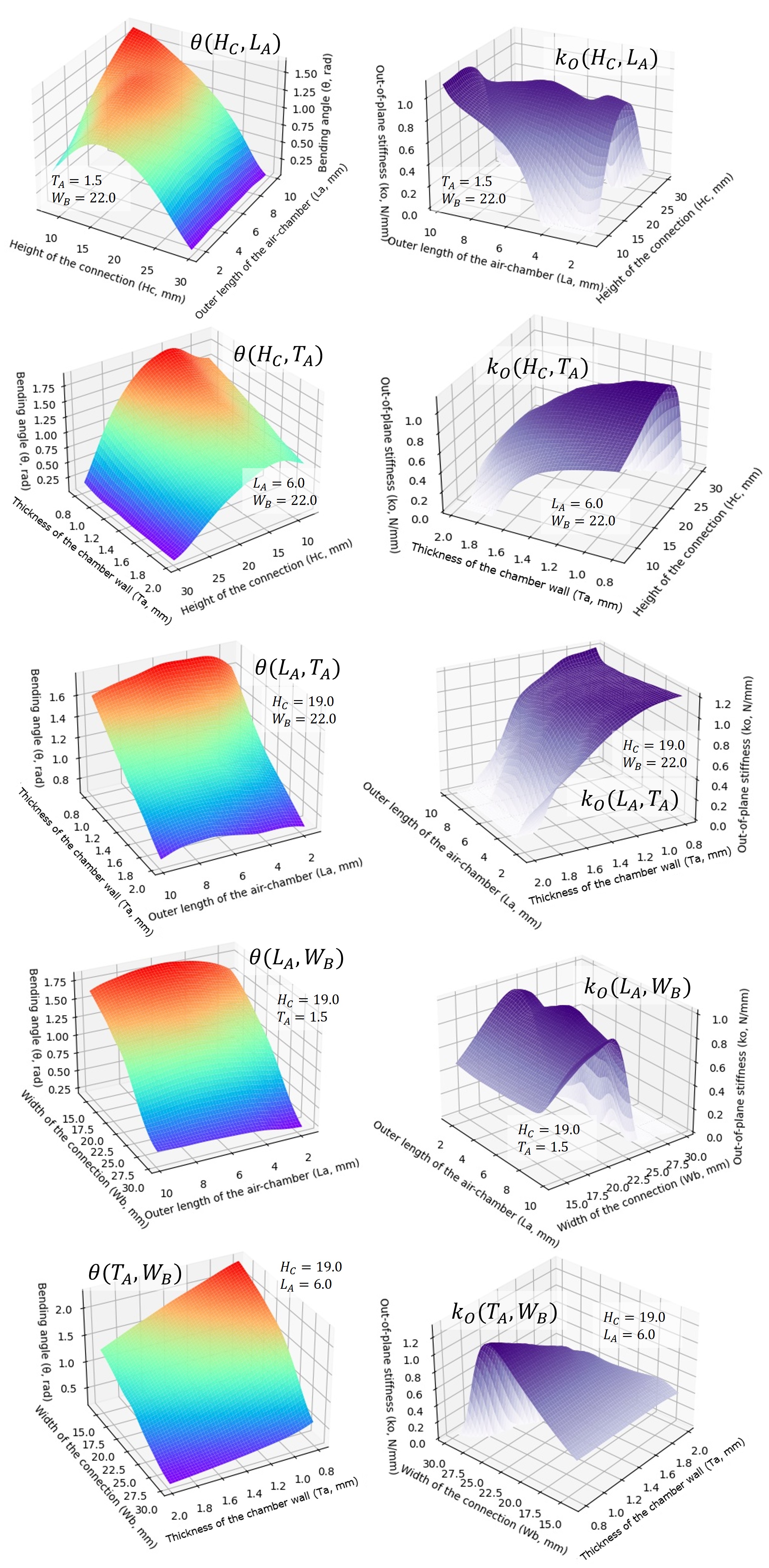}\\
\caption{The response surfaces of $\theta$ and $k_{o}$ respect to different parameters, which \rev{is}{was} determined by our data-driven approach.}
\label{fig:Metamod5}
\end{figure}

A comprehensive visualization of the learned functions for $\Tilde{\theta}(\bm{x})$ and $\Tilde{k}_0(\bm{x})$ can be in Fig.\ref{fig:Metamod5} together with Fig.\ref{fig:Error_Metamod}(b, c).
